\documentclass[conference]{IEEEtran}

\IEEEoverridecommandlockouts
\usepackage{booktabs} % 用于绘制漂亮的表格横线
\usepackage{amsmath}  % 用于数学符号，如 \pm
\usepackage{multirow} % 用于多行单元格
\usepackage{makecell}
\usepackage{xcolor}
\usepackage{colortbl}
\usepackage{cite}
\usepackage{hyperref}
\usepackage{cleveref}
\usepackage{amsmath,amssymb,amsfonts}
\usepackage{algorithmic}
\usepackage{graphicx}
\usepackage{textcomp}
\usepackage{pifont}
\usepackage{xcolor}

\def\BibTeX{{\rm B\kern-.05em{\sc i\kern-.025em b}\kern-.08em
    T\kern-.1667em\lower.7ex\hbox{E}\kern-.125emX}}
\begin{document}

\title{TiS-TSL: Image-Label Supervised Surgical Video Stereo Matching via Time-Switchable Teacher-Student Learning \\
		{}\thanks{$^\dag$: Co-first authors, $^*$: Corresponding authors.}
}
% \author{\IEEEauthorblockN{1\textsuperscript{st} Rui Wang}
% \IEEEauthorblockA{\textit{Wuhan National Laboratory for Optoelectronics} \\
% \textit{Huazhong University of Science and Technology}\\
% Wuhan, China\\
% wwangrui@hust.edu.cn}
% \and
% \IEEEauthorblockN{2\textsuperscript{nd} Ying Zhou}
% \IEEEauthorblockA{\textit{Wuhan National Laboratory for Optoelectronics} \\
% \textit{Huazhong University of Science and Technology}\\
% Wuhan, China\\
% zhouying17@hust.edu.cn}
% \and
% \IEEEauthorblockN{3\textsuperscript{rd} Hao Wang}
% \IEEEauthorblockA{\textit{Wuhan National Laboratory for Optoelectronics} \\
% \textit{Huazhong University of Science and Technology}\\
% Wuhan, China\\
% wanghao020110@hust.edu.cn}
% \and
% \IEEEauthorblockN{4\textsuperscript{th} Wenwei Zhang}
% \IEEEauthorblockA{\textit{Wuhan United Imaging Surgical Co., Ltd.} \\
% Wuhan, China\\
% wenwei.zhang@ui-surgical.com}
% \and
% \IEEEauthorblockN{5\textsuperscript{th} Qiang Li}
% \IEEEauthorblockA{\textit{Wuhan National Laboratory for Optoelectronics} \\
% \textit{Huazhong University of Science and Technology}\\
% Wuhan, China\\
% liqiang8@hust.edu.cn}
% \and
% \IEEEauthorblockN{6\textsuperscript{th} Zhiwei Wang}
% \IEEEauthorblockA{\textit{Wuhan National Laboratory for Optoelectronics} \\
% \textit{Huazhong University of Science and Technology}\\
% Wuhan, China\\
% zwwang@hust.edu.cn}
% }

\author{\IEEEauthorblockN{Rui Wang$^{a,\dag}$, Ying Zhou$^{a,\dag}$, Hao Wang$^{a}$, Wenwei Zhang$^{b}$, Qiang Li$^{a,*}$, Zhiwei Wang$^{a,*}$}
		\IEEEauthorblockA{\textit{$^a$Wuhan National Laboratory for Optoelectronics, Huazhong University of Science and Technology} \\
			\textit{$^b$Wuhan United Imaging Surgical Co., Ltd.}\\
			%Wuhan, China\\
			\{liqiang8, zwwang\}@hust.edu.cn}
		%\and
		%\IEEEauthorblockN{Shiquan He}
		%\IEEEauthorblockA{\textit{Wuhan National Laboratory for Optoelectronics,Huazhong University of Science and Technology} \\
			%	\textit{Huazhong University of Science and Technology}\\
			%	Wuhan, China \\
			%	email address or ORCID}
		%\and
		%\IEEEauthorblockN{Fan Huang}
		%\IEEEauthorblockA{\textit{Wuhan United Imaging Healthcare Surgical Technology Co., Ltd} \\
			%%\textit{}\\
			%Wuhan, China \\
			%email address or ORCID}
		%\and
		%\IEEEauthorblockN{Mei Liu}
		%\IEEEauthorblockA{\textit{Department of Gastroenterology, Tongji Hospital, Tongji Medical College} \\
			%\textit{Huazhong University of Science and Technology}\\
			%Wuhan, China \\
			%email address or ORCID}
		%\and
		%\IEEEauthorblockN{Qiang Li}
		%\IEEEauthorblockA{\textit{Wuhan National Laboratory for Optoelectronics} \\
			%	\textit{Huazhong University of Science and Technology}\\
			%	Wuhan, China \\
			%	liqiang8@hust.edu.cn}
		%\and
		%\IEEEauthorblockN{Zhiwei Wang}
		%\IEEEauthorblockA{\textit{Wuhan National Laboratory for Optoelectronics} \\
			%	\textit{Huazhong University of Science and Technology}\\
			%	Wuhan, China \\
			%	zwwang@hust.edu.cn}
	}

\maketitle

\begin{abstract}
Stereo matching in minimally invasive surgery (MIS) is essential for next-generation navigation and augmented reality. Yet, dense disparity supervision is nearly impossible due to anatomical constraints, typically limiting annotations to only a few image-level labels acquired before the endoscope enters deep body cavities. Teacher-Student Learning (TSL) offers a promising solution by leveraging a teacher trained on sparse labels to generate pseudo labels and associated confidence maps from abundant unlabeled surgical videos. However, existing TSL methods are confined to image-level supervision, providing only spatial confidence and lacking temporal consistency estimation. This absence of spatio-temporal reliability results in unstable disparity predictions and severe flickering artifacts across video frames. To overcome these challenges, we propose TiS-TSL, a novel time-switchable teacher-student learning framework for video stereo matching under minimal supervision. At its core is a unified model that operates in three distinct modes: Image-Prediction (IP), Forward Video-Prediction (FVP), and Backward Video-Prediction (BVP), enabling flexible temporal modeling within a single architecture. Enabled by this unified model, TiS-TSL adopts a two-stage learning strategy. The Image-to-Video (I2V) stage transfers sparse image-level knowledge to initialize temporal modeling. The subsequent Video-to-Video (V2V) stage refines temporal disparity predictions by comparing forward and backward predictions to calculate bidirectional spatio-temporal consistency. This consistency identifies unreliable regions across frames, filters noisy video-level pseudo labels, and enforces temporal coherence. Experimental results on two public datasets demonstrate that TiS-TSL exceeds other image-based state-of-the-arts by improving TEPE and EPE by at least 2.11\% and 4.54\%, respectively.
\end{abstract}


\section{Introduction}
Stereo matching from video is a fundamental task with wide-ranging applications in minimally invasive surgery (MIS), such as dynamic 3D surgical scene reconstruction~\cite{b1}, robotic surgical navigation~\cite{b2}, and augmented/mixed reality guidance~\cite{b3}~\cite{b4}. While extensively studied in natural scenes~\cite{b5}~\cite{b6}, MIS scenarios remain highly challenging, primarily due to the scarcity of high-quality training labels. 

Unlike open and accessible natural environments, MIS procedures are performed in narrow, device-constrained anatomical spaces where integrating additional depth-sensing equipment is often infeasible. As a result, stereo video datasets in surgery are difficult to annotate at scale. For example, the mostly-employed SCARED dataset provides accurate depth maps for only a small number of keyframes, typically the first frame after the endoscope enters the body cavity. This lack of dense supervision makes fully supervised video stereo matching impractical in real-world surgical applications.

To address the problem of limited annotations, recent studies~\cite{b7}~\cite{b8} have explored semi-supervised or self-supervised methods. The semi-supervised method based on Teacher-Student Learning (TSL)~\cite{b9} has been widely used due to its superior performance. A typical strategy involves training an image-based teacher model on a few labeled frames, then using it to generate pseudo labels for the large number of unlabeled frames. A student model is then trained on these pseudo labels, often guided by a confidence estimation module to suppress unreliable regions. This method shows promising results in static image settings, achieving a good trade-off between annotation efficiency and model performance.

Despite these successes, extending semi-supervised frameworks from static stereo images to dynamic stereo videos remains largely underexplored, particularly in endoscopic settings. In our preliminary experiments, directly applying an image-based teacher-student pipeline to stereo video resulted in severe depth flickering artifacts, where depth predictions for the same anatomical region fluctuated abruptly across consecutive frames. We attribute this instability to the absence of temporal consistency in both the pseudo labels and their associated confidence maps. The image-based confidence module is inherently limited to per-frame analysis and cannot detect temporally unstable regions. Consequently, temporally inconsistent supervision persists during training, degrading the continuity of stereo matching over time. 

The core issue stems from a fundamental mismatch between the image-level teacher and the video-level student: the former produces time-independent pseudo labels with spatial-only confidences, while the latter must learn temporally coherent depth representations. This mismatch raises a critical and unresolved question: \emph{How can a small number of labeled frames be leveraged to construct temporally consistent pseudo labels and, crucially, a reliable filtering mechanism for stereo video?} Addressing this challenge is essential for achieving robust and temporally stable stereo matching in real-time surgical environments, where continuity is critical but video-level disparity/depth annotations are typically unavailable.

In this paper, we answer the above question by proposing TiS-TSL, a hierarchical and time-switchable teacher-student learning framework for semi-supervised surgical video stereo matching, requiring only a few image-level disparity labels as supervision. At its core is a time-switchable model that unifies image and video inference within a single architecture. Specifically, the model employs Gated Recurrent Units (GRUs) for both iterative disparity refinement and temporal connection between adjacent frames. This design enables the model to operate in three modes: Image-Prediction (IP), Forward Video-Prediction (FVP), and Backward Video-Prediction (BVP), depending on how the temporal connections are configured.

Built upon this time-switchable model, we design a two-stage teacher-student framework, comprising Image-to-Video (I2V) and Video-to-Video (V2V) stages, which sequentially enhance video stereo matching from limited frame-wise supervision. The I2V stage transfers image-level knowledge to the video domain by using a teacher in image mode to generate pseudo labels that supervise a student in video mode, initializing its temporal modeling capability. The V2V stage further refines this by having the teacher in forward/ backward video modes reason bidirectionally through time. By comparing predictions from forward and backward passes, the model learns to identify its own inconsistencies. This yields a \emph{spatio-temporal} confidence map that filters noisy supervision and enables the student to focus on temporally stable signals, significantly improving disparity continuity and robustness. 

In summary, our contributions are listed as follows:
% \vspace{-0.8em}
\begin{itemize}
\item We propose a novel time-switchable video stereo matching model that extends conventional image-based GRU architectures with minimal modifications to enable flexible temporal modeling. This unified design seamlessly integrates image and video inference, enhancing training efficiency, prediction accuracy, and temporal coherence, forming the backbone of our hierarchical semi-supervised teacher-student framework.
\item We introduce TiS-TSL, a time-switchable teacher-student learning, consisting of I2V and V2V stages. TiS-TSL leverages a spatio-temporal confidence filtering mechanism based on soft consistency, which robustly identifies and suppresses unreliable pseudo label regions across both spatial and temporal domains, greatly improving the reliability and temporal consistency of video stereo matching.
\item We conduct experiments on two public endoscopic stereo datasets, and demonstrate that TiS-TSL achieves state-of-the-art results on public endoscopic stereo datasets. Compared to image-based state-of-the-arts, TiS-TSL reduces TEPE and EPE by over 2.11\% and over 4.54\%, respectively. Crucially, it requires supervision from only a single labeled frame per video, yet produces robust, temporally consistent disparity maps that effectively eliminate the flickering artifacts.
\end{itemize}
\section{Related Work}
\subsection{Image-Based Stereo Matching}
Image-based stereo matching is a classic problem in computer vision and has been extensively studied in natural scenes. Recently, iterative optimization-based methods~\cite{b10} -~\cite{b12} have achieved leading performance in accuracy and robustness, becoming the mainstream paradigm. These methods typically begin with an initialized disparity map, then iteratively refine the disparity through a cascading process. Inspired by the optical flow method RAFT~\cite{b13}, RAFT-Stereo~\cite{b10} pioneers this approach for stereo matching, introducing multi-layer GRUs to expand the receptive field. Unlike RAFT-Stereo, which initializes disparity to zero, IGEV-Stereo~\cite{b11} regresses a combined geometric encoding volume for a more accurate initial disparity, accelerating convergence. Selective-Stereo~\cite{b12} proposes a Selective Recurrent Unit (SRU) to selectively process high-frequency and low-frequency regions. Despite impressive performance, applying these methods directly to video disparity estimation often causes severe temporal flickering artifacts due to frame-wise independence. 
\subsection{Video-Based Stereo Matching}
Video-based stereo matching has been a relatively less explored area. Recently, some methods have addressed stereo matching in videos to improve the temporal consistency of disparity estimation. CODD~\cite{b14} predicts SE(3) transformations per pixel to align previous-frame disparity to the current frame and performs information fusion. DynamicStereo~\cite{b5} uses Transformer~\cite{b15} to capture inter-frame temporal continuity and optimize predictions, maintaining high consistency in complex dynamic scenes. BiDAStereo~\cite{b6} designs a triple-frame correlation layer, warping features from the previous and subsequent frames towards the center frame using optical flow to ensure feature alignment. While demonstrating excellent temporal consistency, these methods rely heavily on large-scale video annotations, hindering their application in MIS scenarios with limited labeled data.

\subsection{Semi-supervised Stereo Matching}
Unlike natural scenes, annotated samples in MIS scenarios are difficult to obtain. To reduce label dependency, semi-supervised training approaches can be employed. These methods combine the advantages of self-supervised and fully supervised methods. Among these, Teacher-Student Learning (TSL) is widely applied. Shi~\emph{et al.}~\cite{b7} proposes an improved teacher-student model to effectively guide the student, additionally training a spatial confidence network to suppress erroneous information in teacher-generated pseudo labels. The subsequent work BiSD~\cite{b16} builds on a similar idea and further proposes a dual-branch teacher-student model, where each branch can provide confidence, better suppressing erroneous pseudo labels and leading to steadier training convergence. However, current stereo matching confidence mechanisms only consider the spatial dimension. How to extend it to semi-supervised video stereo matching still remains underexplored.

\begin{figure*}[htbp] % 'figure*' 表示这是一个跨两栏的浮动图片环境
\centering % 使图片在环境中水平居中
\includegraphics[width=0.87\textwidth]{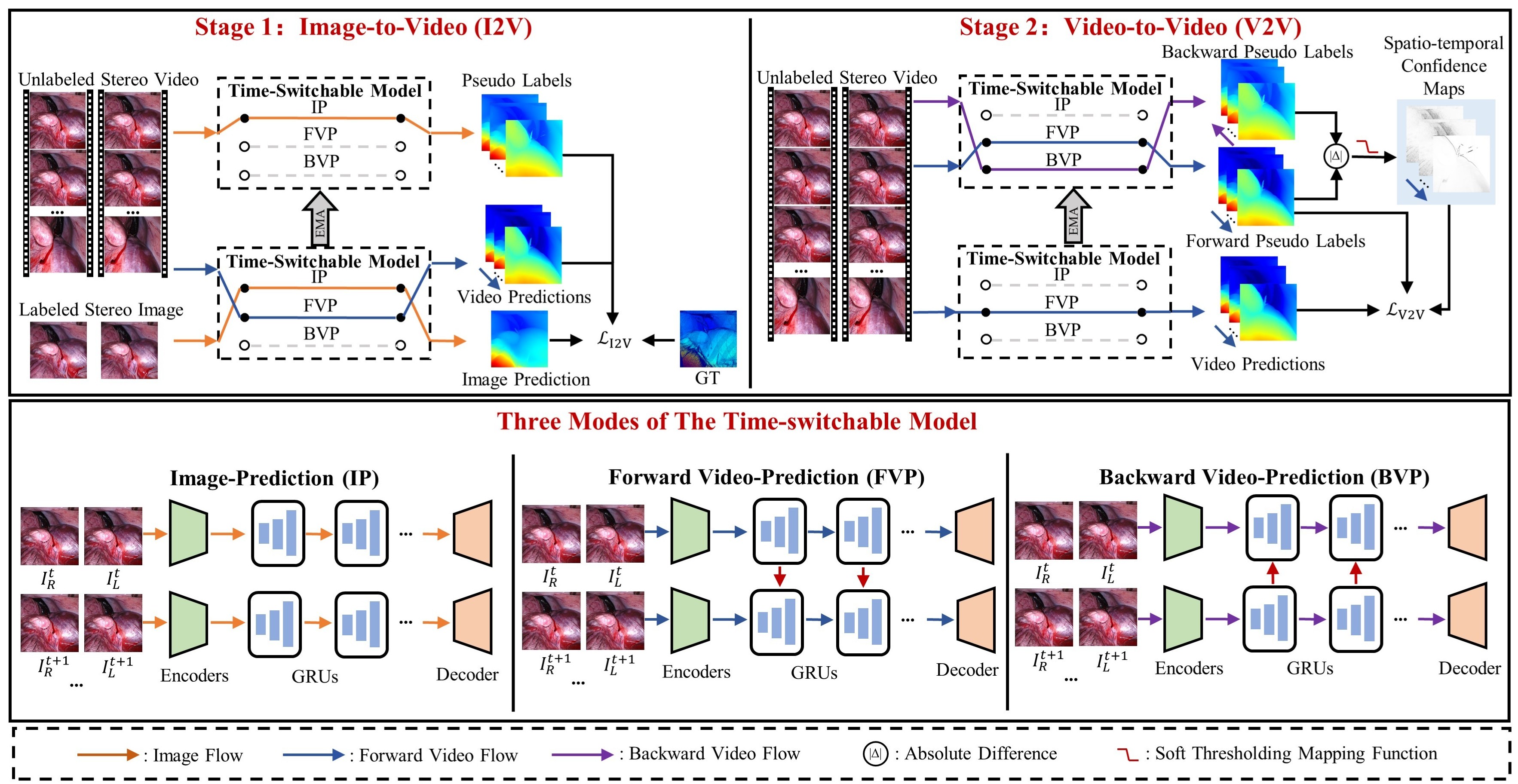} 

\caption{The overview of our proposed TiS-TSL. At its core is a time-switchable model, which operates in three
distinct modes based on the flow patterns of the hidden states within the GRUs, i.e., Image-Prediction (IP), Forward Video-Prediction (FVP), and Backward Video-Prediction (BVP). Enabled by this model, TiS-TSL comprises two stages, i.e., Image-to-Video (I2V) stage and Video-to-Video (V2V) stage. The I2V stage aims to pretrain spatial components using labeled images while initializing temporal modeling on unlabeled video via pseudo-supervision. The subsequent V2V stage applies bidirectional prediction consistency to filter out unreliable regions in the pseudo labels, thereby explicitly reinforcing temporal consistency.} % 图片标题
\label{fig:OverView} % 图片标签，用于在文中引用

\end{figure*}
\section{Methods}
The overview of TiS-TSL is illustrated in Fig.~\ref{fig:OverView}. At its core is a time-switchable stereo model, trained through a two-stage paradigm: Image-to-Video (I2V) and Video-to-Video (V2V). We detail each component below.

\subsection{Time-Switchable Model}\label{AA}
To support both images and videos, we design a time-switchable stereo model that operates in three modes: Image-Prediction (IP), Forward Video-Prediction (FVP), and Backward Video-Prediction (BVP). Our model builds upon IGEV-Stereo~\cite{b11}, and consists of \textbf{(i) Feature Extraction:} Given a stereo pair $\{I_l, I_r\} \in \mathbb{R}^{3 \times H \times W}$, a Feature Encoder extracts hierarchical features at four scales: 1/32, 1/16, 1/8, and 1/4. Low-resolution features are progressively upsampled, concatenated, and fused to yield rich 1/4-scale representations.
A geometry volume $G \in \mathbb{R}^{D \times H/4 \times W/4}$ is constructed using group-wise correlation over a disparity range $D$. Simultaneously, a Context Encoder processes only the left image $I_l$ to generate context features $F_c \in \mathbb{R}^{C \times H/4 \times W/4}$.
\textbf{(ii) Disparity Refinement:} A light 3D CNN first converts $G$ to an initial disparity map $d^0 \in \mathbb{R}^{H/4 \times W/4}$, which is then refined over $N$ iterations to produce finer and finer estimation. The core of the refinement process is a GRU that integrates context and geometry cues into the hidden state $h^i \in \mathbb{R}^{C \times H/4 \times W/4}$. The update is defined as:
\begin{equation}
\left\{
\begin{aligned}
&d^i = d^{i-1} + \Delta d^{i-1} \\
&\Delta d^{i} = \mathtt{conv}(h^{i})\\
&h^{i} = \mathtt{GRU}(x^i,h^{i-1})
\end{aligned}
\right.
\end{equation}

Here, $i=1,\dots,N$, and $x^i$ is the fused input at iteration $i$, encoding disparity-aware features. The hidden state is initialized as $h^0 = \mathtt{tanh}(F_c)$. To accommodate both image and video inputs, we propose a unified GRU update scheme with switchable $h^i$ generation logic. The key idea is to tailor $h^i$ to either static or temporally varying inputs using one of three modes.

% \begin{figure}[htbp] 
% \centering % 使图片在环境中水平居中
% \includegraphics[width=0.48\textwidth]{fig2.jpg} 

% \caption{Iteration disparity refinement process in IP mode. The initial disparity and initial hidden state are obtained from the geometry volume and context feature, respectively. After $N$ iterations by GRUs, the final refined disparity can be obtained.} 
% \label{fig:GRUs} % 图片标签，用于在文中引用

% \end{figure}

\subsubsection{Image-Prediction (IP) mode} 
In the static image setting, our model performs iterative disparity refinement by fusing geometry, disparity, and context cues without considering the temporally adjacent frames. At iteration $i$ in Disparity Refinement, we extract geometry feature map $F_g\in \mathbb{R}^{C \times H/4 \times W/4}$ from the volume $G$ by using the values in the previous disparity map $d^{i-1}$ as indexes to index the 3D cost volume $G$ along the disparity dimension.
We then encode both $F_g$ and $d^{i-1}$ with convolutional layers $\mathtt{conv}_g$ and $\mathtt{conv}_d$, and combine them with context features to form the GRU input:
\begin{equation}
\small
x^i = \mathtt{concat}(\mathtt{conv}_g(F_g), \mathtt{conv}_d(d^{i-1}), d^{i-1}, \mathtt{ReLU}(F_c)).
\label{eq}
\end{equation}

The GRU maintains a hidden state $h^i$ and integrates current features $x^i$ with prior information. This allows the model to correct errors and aggregate context across refinement steps iteratively. The GRU update follows:
\begin{align}
z^i &= \sigma(\mathtt{conv}_z(\mathtt{concat}(h^{i-1}, x^i))) \nonumber \\
r^i &= \sigma(\mathtt{conv}_r(\mathtt{concat}(h^{i-1}, x^i))) \nonumber  \\
\tilde{h}^i &= \mathtt{tanh}(\mathtt{conv}_h(\mathtt{concat}(r^i\odot h^{i-1}, x^i))) \label{eq:combined_equations}\\
h^i &= (1-z^i)\odot h^{i-1} + z^i \odot \tilde{h}^i \nonumber 
\end{align}

Here, $\odot$ denotes element-wise product, $\sigma$ denotes sigmoid. $z^i$ and $r^i$ are the update and reset gates, respectively, which control the flow of information from past states and current inputs. This recurrent mechanism allows the model to refine disparities in a coarse-to-fine manner, capturing both local details and long-range dependencies.

\subsubsection{Forward Video-Prediction (FVP) mode} To enable temporal propagation of disparity information, we extend the intra-frame propagation (IP) mode by incorporating hidden state interactions across adjacent frames.
To be clearer, we use a temporal index $t$ ($t=1,..., T$) to distinguish features at different time steps.
For the $t$-th frame at $i$-th GRU iteration, the input hidden state $h_t^{i-1}$ is obtained by fusing itself with the corresponding hidden state $h_{t-1}^{i-1}$ from the preceding frame:
\begin{equation}
h^{i-1}_{t} = \mathtt{conv}_f(\mathtt{cam}(\mathtt{concat}(h^{i-1}_t, h^{i-1}_{t-1})))\label{eq}
\end{equation}

Here, $\mathtt{cam}$ denotes the channel attention module~\cite{b17} to emphasize informative channels across time, and $\mathtt{conv}_f$ projects the resulting feature to fit the original dimension. The new temporally-related hidden state $h^{i-1}_{t}$ is then used in the GRU update to generate $h_t^i$, following the same update rule as in IP mode.
If $t=1$, i.e., the first frame in the sequence, no preceding temporal context is available. In this case, the model skips fusion and performs intra-frame propagation only.

\subsubsection{Backward Video-Prediction (BVP) mode}BVP mode mirrors the forward setup but propagates information in the reverse temporal direction. For the $t$-th frame at $i$-th GRU iteration, the hidden state $h_t^{i-1}$ is obtained by fusing itself with $h_{t+1}^{i-1}$ from the next frame:
\begin{equation}
h^{i-1}_{t} = \mathtt{conv}_f(\mathtt{cam}(\mathtt{concat}(h^{i-1}_t, h^{i-1}_{t+1})))
\end{equation}

This enables backward propagation of temporal cues, complementing the forward pathway. Similarly, if $t$ corresponds to the last frame in the sequence, the model defaults to IP mode.

\subsection{Image-to-Video (I2V) Stage Initializes Temporal Modeling}
We denote our time-switchable model as a function $\mathtt{TiS}$, supporting three inference modes: image prediction ($\mathtt{TiS}_{\mathtt{IP}}$), forward video prediction ($\mathtt{TiS}_{\mathtt{FVP}}$), and backward video prediction ($\mathtt{TiS}_{\mathtt{BVP}}$). The temporal fusion module (see Eq. (4) and Eq. (5)) is active only in $\mathtt{FVP}$ and $\mathtt{BVP}$ modes to incorporate temporal context.
Let $\mathcal{I} \in \mathbb{R}^{6 \times H \times W}$ denote a labeled stereo image pair (left and right views) with ground-truth disparity $\mathcal{D} \in \mathbb{R}^{H \times W}$. An unlabeled stereo video clip with $T$ consecutive frames is represented as $\mathcal{V} = \{ \mathcal{I}_t \}_{t=1}^{T}$.

The I2V stage aims to pre-train spatial components using labeled images while initializing temporal modeling on unlabeled video via pseudo-supervision.
To this end, we adopt a teacher-student learning framework where both networks share architecture but maintain separate parameters, denoted as $\mathtt{TiS}(\cdot;\theta_{\text{tea}})$ and $\mathtt{TiS}(\cdot;\theta_{\text{stu}})$, respectively. The final predicted disparity map $d^N$ (at $1/4$ resolution) is upsampled, and L1 loss is applied following IGEV-Stereo~\cite{b11}. 

The overall I2V objective is defined as:
\begin{align}
\small
\mathcal{L}_{\text{I2V}} = &\mathbb{E}_{\mathcal{I}}[ \| \mathtt{TiS}_{\mathtt{IP}}(\mathcal{I}; \theta_{\text{stu}}) - \mathcal{D} \|_1] + \\ \nonumber
&\mathbb{E}_{\mathcal{V}} [\sum_{t=1}^{T} \| \mathtt{TiS}_{\mathtt{FVP}}(\mathcal{I}_t; \theta_{\text{stu}}) - \mathtt{TiS}_{\mathtt{IP}}(\mathcal{I}_t; \theta_{\text{tea}}) \|_1].
\label{eq:7}
\end{align}
where $\mathbb{E}_{\mathcal{I}}[\cdot]$ and $\mathbb{E}_{\mathcal{V}}[\cdot]$ denote the expectations over labeled stereo images and unlabeled video clips, respectively.

The first term in Eq.~(6) supervises the student in $\mathtt{IP}$ mode using static images and their ground-truth disparities, enabling learning of spatial and refinement modules. The second term in Eq.~(6) uses the teacher (in $\mathtt{IP}$ mode) to generate frame-wise pseudo labels of unlabeled videos that guide the student (in $\mathtt{FVP}$ mode) to learn the temporal relationship.

$\mathcal{L}_{\text{I2V}}$ updates only the student parameters $\theta_{\text{stu}}$, while the teacher weights $\theta_{\text{tea}}$ are updated via exponential moving average (EMA):
\begin{equation}
\theta_{\text{tea}} = \alpha \theta_{\text{tea}} + (1 - \alpha) \theta_{\text{stu}}.
\end{equation}
where $\alpha$ is a momentum coefficient. This EMA update stabilizes training by smoothing the teacher’s predictions and avoiding collapse during pseudo-supervision.

\subsection{Video-to-Video (V2V) Stage Boosts Temporal Consistency}
Although the I2V stage equips $\mathtt{TiS}$ with preliminary temporal modeling, it relies on frame-level pseudo labels without enforcing cross-frame consistency, often leading to temporal artifacts such as flickering. To address this, the V2V stage explicitly enhances temporal coherence by supervising the model with sequence-level consistency signals in a teacher-student framework.
A key challenge is that the temporal predictions from $\mathtt{TiS}$ remain noisy and unreliable, particularly in occluded or low-texture regions. This limits the effectiveness of direct self-training with video sequences.

\begin{table*}[t]
\centering
\caption{Quantitative comparison with image-based methods on SCARED and Hamlyn. The best results are marked in bold. The paired $p$-values between TiS-TSL and others are all less than 0.05.}
\label{tab:comparison}
\resizebox{\textwidth}{!}{%
\begin{tabular}{lcccccccccccc}
\toprule
\multirow{2}{*}{Method} & \multicolumn{6}{c}{\textbf{SCARED}} & \multicolumn{6}{c}{\textbf{Hamlyn}} \\
\cmidrule(lr){2-7} \cmidrule(lr){8-13} % cmidrule 范围不变
 & TEPE ↓ & $\delta_t^{3px}$ ↓ & TEPE$_r$ ↓ & $\delta_t^{100\%}$ ↓ & EPE ↓ & $\delta^{3px}$ ↓ 
 & TEPE ↓ & $\delta_t^{3px}$ ↓ & TEPE$_r$ ↓ & $\delta_t^{100\%}$ ↓ & EPE ↓ & $\delta^{3px}$ ↓ \\
\midrule

PSMNet & 1.15 & 4.10 & 8.76 & 60.30 & 2.35 & 18.48 
       & 0.70 & 2.14 & 0.87 & 32.41 & 1.36 & 6.01\\
GwcNet & 1.12 & 3.89 & 8.27 & 58.78 & 2.36 & 17.50
       & 0.69 & 2.15 & 0.86 & 31.40 & 1.25 & 5.38 \\
ACVNet & 1.14 & 3.89 & 8.71 & 58.04 & 2.30 & 17.27
       & 0.69 & 2.13 & 0.87 & 31.70 & 1.28 & 5.35\\
IGEV-Stereo & 1.01 & 3.41 & 6.40 & 56.28 & 2.06& 16.87
           & 0.68 & 2.11& 0.87 & 31.34 & 1.25 & 5.21 \\
Selective-IGEV & 1.04 & 3.43 & 7.01& 56.92 & 2.13 & 16.62
              & 0.68 & 2.09 & 0.86 & 31.98 & 1.26& 5.35 \\
FoundationStereo & 0.99 & 3.30 & 6.30 & 56.34 & 2.03& 16.52 
                & 0.68 & 2.05 & 0.86 & 31.10 & 1.25& 5.33\\
BiSD & 0.95 & 3.04 & 5.91 & 56.84 & 1.98 & 16.47
             & 0.68 & 2.10 & 0.86 & 32.21 & 1.26& 5.56 \\
TiS-TSL (Ours) & \textbf{0.93} & \textbf{2.72} & \textbf{4.99} & \textbf{55.03} & \textbf{1.89} & \textbf{14.38} 
     & \textbf{0.66} & \textbf{1.95} & \textbf{0.85} & \textbf{30.85} & \textbf{1.17} & \textbf{4.58} \\
\bottomrule
\end{tabular}%
}
\end{table*}

\begin{figure*}[htbp] 
\centering % 使图片在环境中水平居中
\includegraphics[width=0.85\textwidth]{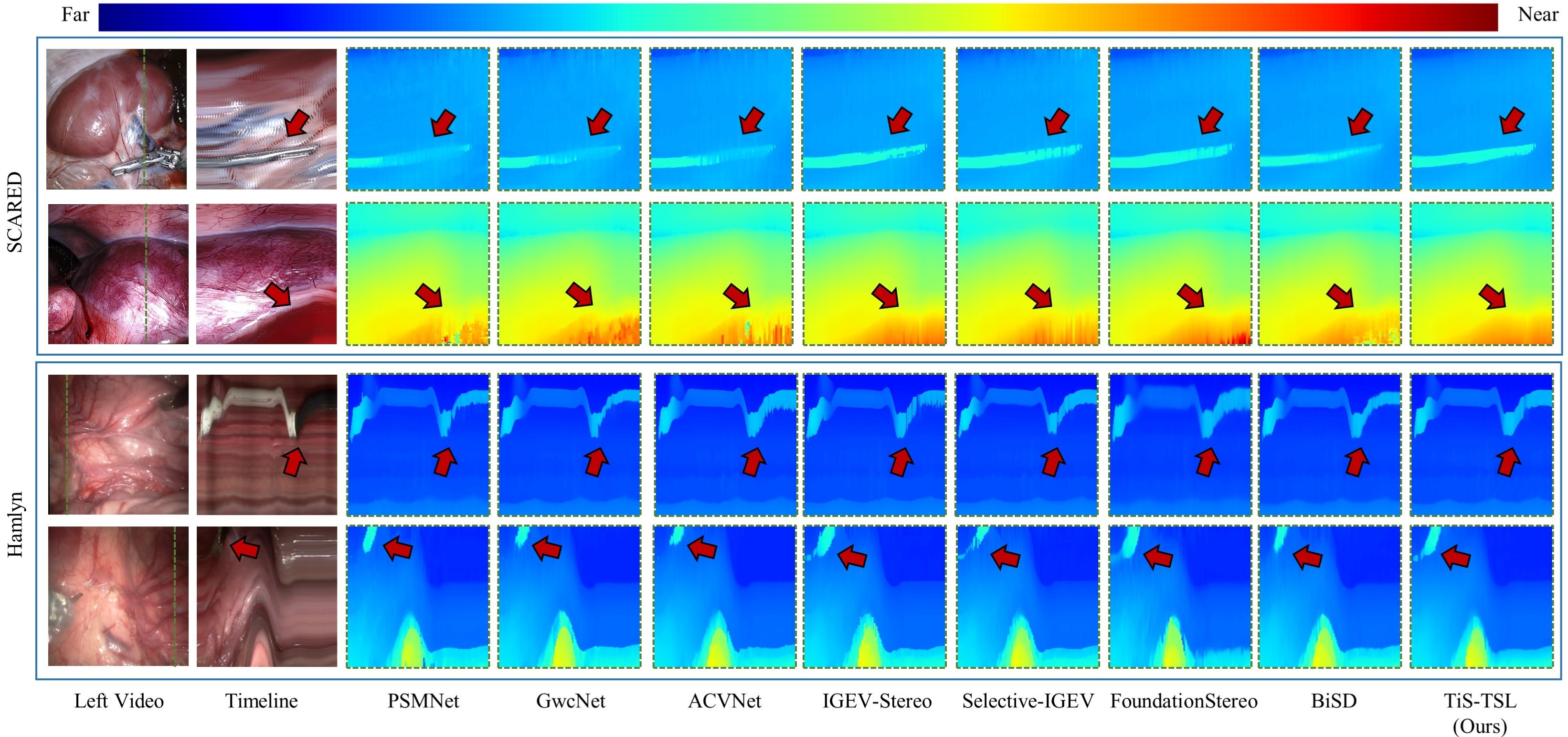} 

\caption{Qualitative comparisons with image-based methods on SCARED and Hamlyn videos. The second column represents the temporal image profiles obtained by slicing the images along the timeline at green-line positions. The subsequent columns represent the corresponding disparity profiles. The red arrows indicate highlight regions. OpenCV Jet Colormap is used for visualization.} 
\label{fig:timeline} % 图片标签，用于在文中引用

\end{figure*}

To tackle this, we propose a Spatio-Temporal Confidence Filtering Mechanism (ST-CFM) to quantify the temporal consistency between forward and backward predictions and use it as a proxy for pseudo label reliability, thereby enabling selective supervision.
Specifically, for a video $\mathcal{V} = \{ \mathcal{I}_t \}_{t=1}^{T}$, we generate bidirectional disparity predictions using the teacher model operating in both $\mathtt{FVP}$ and $\mathtt{BVP}$ modes:
\begin{equation}
\left\{
\begin{aligned}
&\hat{\mathcal{D}}_t^f=\mathtt{TiS}_{\mathtt{FVP}}(\mathcal{I}_t;\theta_{\text{tea}}) \\
&\hat{\mathcal{D}}_t^b=\mathtt{TiS}_{\mathtt{BVP}}(\mathcal{I}_t;\theta_{\text{tea}})
\end{aligned}
\right.
\end{equation}

Intuitively, if the model captures consistent temporal dynamics, the forward and backward predictions should match closely for the same frame. Therefore, we define a soft confidence map $W_t$ for each frame by transforming the absolute disparity difference into a reliability weight via a sigmoid-shaped soft thresholding mapping function \cite{b18}:
\begin{equation}
W_t = \frac{1}{1 + e^{\epsilon (|\hat{\mathcal{D}}_t^f - \hat{\mathcal{D}}_t^b|- \tau)}} \label{eq:soft}
\end{equation}
where $\epsilon$ is a scale factor controlling the sharpness of the transition, and $\tau$ is a soft threshold that determines tolerance for disparity disagreement.
The spatio-temporal confidence map selectively emphasizes temporally consistent regions while attenuating noisy supervision, effectively guiding the student to focus on stable spatio-temporal patterns.

The objective of the V2V stage can be formulated as a confidence-weighted temporal distillation:
\begin{align}
\small
\mathcal{L}_{\text{V2V}} = \mathbb{E}_{\mathcal{V}} [\sum_{t=1}^{T} \| &W_t\odot\mathtt{TiS}_{\mathtt{FVP}}(\mathcal{I}_t; \theta_{\text{stu}}) \\ \nonumber
-&W_t\odot\mathtt{TiS}_{\mathtt{FVP}}(\mathcal{I}_t; \theta_{\text{tea}}) \|_1].
\label{eq}
\end{align}

This formulation ensures that temporal learning focuses on spatio-temporally reliable regions, thereby reducing flickering and enhancing prediction stability across frames. 
Likewise, $\mathcal{L}_{\text{V2V}}$ updates only the student parameters $\theta_{\text{stu}}$, while the teacher weights $\theta_{\text{tea}}$ are updated via EMA as formulated in Eq. (7).

\section{EXPERIMENTAL SETTINGS}
\subsection{Implementation Details}

We train TiS-TSL on two NVIDIA GeForce RTX 4090 GPUs. We select AdamW~\cite{b19} as the optimizer and use the one-cycle learning rate schedule with a learning rate of 0.0002. We conduct the same data augmentation as~\cite{b11} and set the batch size to 4. Following~\cite{b11}, we pre-train our time-switchable stereo model on Scene Flow for 200k steps. In both FVP and BVP modes, we set the sequence length to 4 frames. The number of GRU iterations is set to 22 for training and reduced to 12 for efficient inference. For spatio-temporal confidence filtering, the $\epsilon$ and $\tau$ are set to be 10 and 1, respectively.

\subsection{Datasets and Metrics}
We conduct comprehensive experiments on two endoscopic datasets, including SCARED and Hamlyn~\cite{b20}. 

\subsubsection{SCARED} SCARED is from one of the MICCAI 2019 challenges and contains 7 subjects for training. Each subject corresponds to five videos recorded by a binocular endoscope, with a resolution of 1024×1280. Each video contains one keyframe. The ground-truth (GT) depth maps of keyframes were obtained via structured light, while the depth maps of other frames were interpolated from poses recorded by the robotic arm.

Due to calibration parameter errors, the depth annotations for the keyframes are only completely accurate in five of the seven subjects (excluding the fourth and fifth ones). Additionally, because of the accumulated error in the poses recorded by the robotic arm over time, the video frames and their corresponding depth maps become temporally misaligned. Therefore, during the training phase, we utilize only the $5\times5=25$ image-level GT depth maps as labeled data, while treating all remaining frames as unlabeled data.

\subsubsection{Hamlyn} Hamlyn is a large dataset of various-resolution endoscopic videos obtained using Da Vinci surgical robots. Recasens \textit{et al}.~\cite{b20} provided a rectified version and generated GT depth maps using the \textit{Libelas} software. Videos with a resolution of $288 \times 770$ are excluded due to the poor quality of their GT depth maps. We use the remaining 8 videos for evaluation.

Following the protocol in\cite{b14}, we employ Temporal End-Point Disparity Error (TEPE), threshold metric of 3px for TEPE ($\delta^{3px}_t$), relative error (TEPE$_r$), and threshold metrics of 100\% ($\delta^{100\%}_t$) for TEPE$_r$ to evaluate the temporal consistency. In addition, we employ end-point disparity error (EPE) and threshold metric of 3px ($\delta^{3px}$) to evaluate the accuracy. 

\section{RESULTS AND DISCUSSIONS}
\subsection{Comparison with Image-based State-of-the-arts} 
\label{sec:CIB}
 We compare TiS-TSL with 7 image-based state-of-the-art (SOTA) stereo matching methods, i.e., PSMNet~\cite{b21}, GwcNet~\cite{b22}, ACVNet~\cite{b23}, IGEV-Stereo~\cite{b11}, Selective-Stereo~\cite{b12}, FoundationStereo~\cite{b24}, BiSD~\cite{b16}. Note that BiSD~\cite{b16}, like TiS-TSL, is also a semi-supervised method, while the remaining methods are fully supervised. We evaluate their performance and make
comparisons using their released codes.

\subsubsection{Evaluation on SCARED} Following BiSD~\cite{b16}, we conduct 5-fold cross-validation experiments on the five training subjects of SCARED, i.e., leaving one subject for testing and using the remaining four subjects for training. Due to the accumulated error mentioned before, we only use the first 60 frames (approximately 2 seconds) of each video sequence.
 
The left part of Table \ref{tab:comparison} lists the comparison results between our method and other SOTA methods on SCARED. The result is the average indicator of all subjects. For all metrics, lower values indicate better performance.

It can be seen that semi-supervised methods ($7^{th}$ and $8^{th}$ rows) generally outperform fully supervised methods ($1^{st}$ to $6^{th}$ rows), which can be attributed to the utilization of unlabeled data. Unlike image-based methods that only exploit image-level information, TiS-TSL leverages both image-level and video-level information by switching among three distinct modes. Moreover, TiS-TSL improves the efficiency of pseudo label supervision via the ST-CFM. Compared to second-best BiSD, TiS-TSL improves TEPE and EPE by 2.11\% and 4.54\%, respectively. 

The top two rows of Fig. \ref{fig:timeline} visualize the two video samples on SCARED. The second column represents the temporal image profiles obtained by slicing the images along the timeline at green-line positions. The subsequent columns represent the corresponding disparity profiles. As illustrated in Fig. \ref{fig:timeline}, the disparity profiles estimated by our method are most consistent with the original left videos, particularly in challenging complex contour regions of surgical instruments (e.g., the first row) and in low-texture regions with blood (e.g., the second row). This strongly demonstrates that our method achieves the best temporal consistency.

\begin{figure}[htbp] 
\centering % 使图片在环境中水平居中
\includegraphics[width=0.48\textwidth]{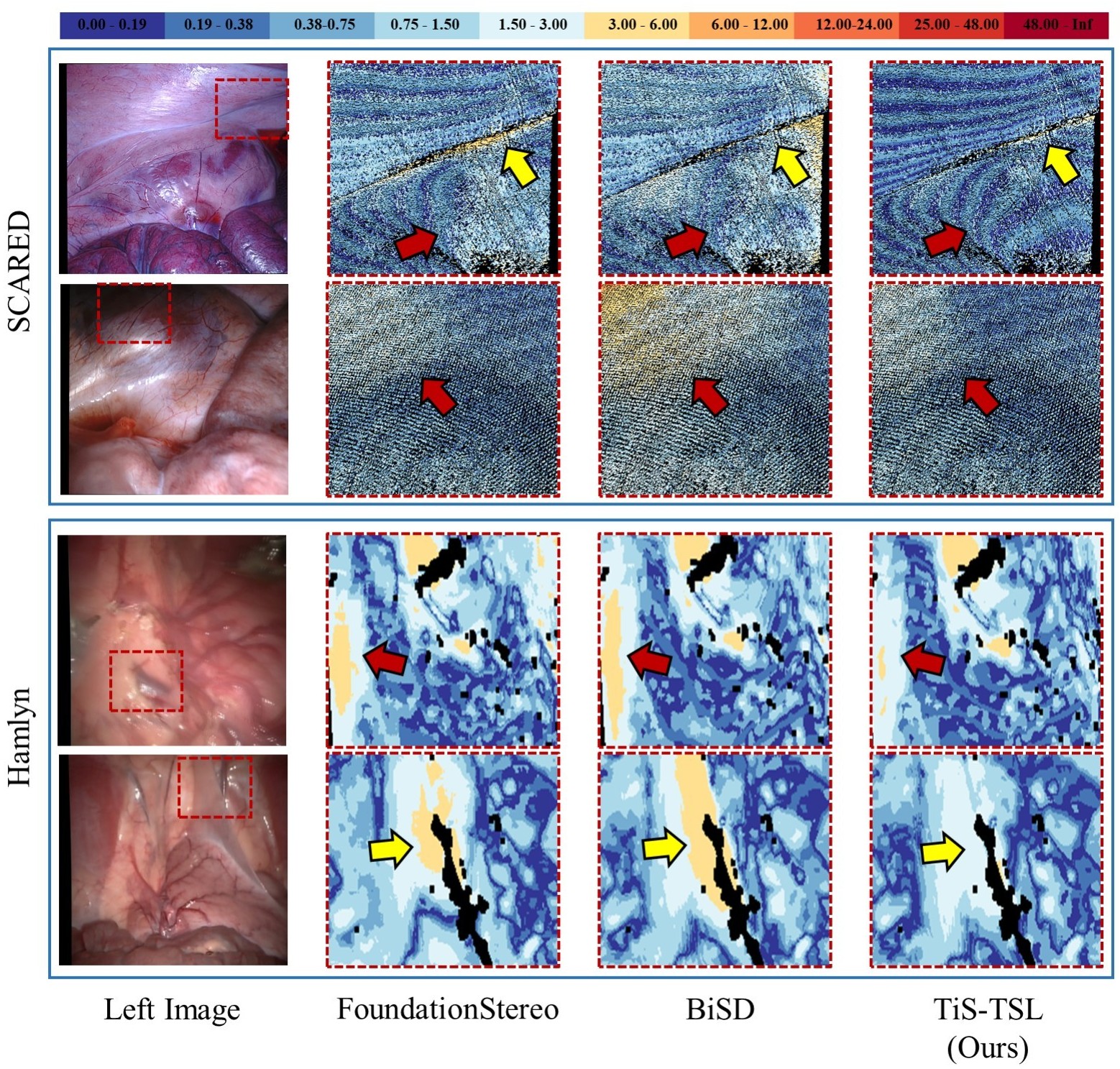} 

\caption{Error maps of predicted disparities on SCARED and Hamlyn. Regions indicated by red arrows in the figure represent flat areas, while those indicated by yellow arrows represent boundary areas. The black regions represent invalid pixels in the GT disparity.} 
\label{fig:Error map} % 图片标签，用于在文中引用

\end{figure}

We further analyze accuracy by comparing TiS-TSL with two representative image-based methods, i.e., FoundationStereo, and BiSD. We obtain the error map by calculating the absolute distance between the GT and predicted disparity for each pixel. The black regions represent invalid pixels in the GT. As illustrated in the top two rows of Fig. \ref{fig:Error map}, our method achieves the most accurate disparity prediction in both organ boundary areas (indicated by yellow arrows) and flat areas (indicated by red arrows).

% \begin{figure}[htbp] 
% \centering % 使图片在环境中水平居中
% \includegraphics[width=0.48\textwidth]{fig4.jpg} 

% \caption{Qualitative comparisons on SCARED videos. The second column represents the temporal image profile obtained by slicing the images along the timeline within a 10-pixel green box. The subsequent columns represent the corresponding disparity profiles. Our TiS-TSL achieves the best temporal consistency in disparity estimation.} 
% \label{fig:timeline} % 图片标签，用于在文中引用

% \end{figure}

\subsubsection{Generalization on Hamlyn}To evaluate the generalizability of methods, we train all models using all five subjects of SCARED and directly test them on Hamlyn. For evaluation, we utilize the first 500 frames from each video. The comparison results are listed in the right part of Table \ref{tab:comparison}. Under such conditions, BiSD demonstrates weaker generalization performance, while TiS-TSL still achieves the best performance. Compared to the second-best FoundationStereo, our method reduces TEPE and EPE by 2.94\% ($p$ $<$ 0.005) and 6.40\% ($p$ $<$ 0.001), respectively. 

The lower two rows of Fig. \ref{fig:timeline} and the lower two rows of Fig. \ref{fig:Error map} visualize two video samples and two image samples on Hamlyn, respectively. As can be seen, compared to other methods, TiS-TSL still delivers the most accurate and temporally consistent disparity estimation, revealing its satisfactory generalization.

\begin{table}[htbp] % 使用 'table' 环境确保表格占据单栏宽度
\centering % 居中表格
\caption{Quantitative comparison with video-based methods on SCARED. All metrics are for 1024×1280 inputs on a single NVIDIA GeForce RTX 4090 GPU. The best results are marked in bold. The paired $p$-values between TiS-TSL and others are all less than 0.05.}
\label{tab:video_comparison} % 表格标签，用于引用

\footnotesize % 减小字体大小，以适应单栏表格

% 列格式定义：
% l: Method 列左对齐
% c: 后面所有指标列居中对齐
\begin{tabular}{
>{\raggedright\arraybackslash}p{2.05cm}
*{1}{>{\centering\arraybackslash}m{1cm}} 
*{1}{>{\centering\arraybackslash}m{0.55cm}}
*{1}{>{\centering\arraybackslash}m{1.1cm}}
*{1}{>{\centering\arraybackslash}m{0.75cm}}
*{1}{>{\centering\arraybackslash}m{0.9cm}}}
    \toprule % 表格顶部粗线
    Method & TEPE $\downarrow$ & $\delta_t^{3px}$$\downarrow$ & TEPEr $\downarrow$ & EPE $\downarrow$& \shortstack{Run-time \\ (ms) $\downarrow$}  \\
    \midrule % 表格中间线

    I2V+DynamicStereo & \textbf{0.93} & 2.86 & 5.11 & 1.97 & 848 \\
    I2V+BiDAStereo & 0.95 & \textbf{2.69} & 5.16 & 1.99 & 1593 \\
    I2V+V2V (Ours) & \textbf{0.93} & 2.78 &  \textbf{4.99} & \textbf{1.89} & \textbf{403} \\
    \bottomrule % 表格底部粗线
\end{tabular}
\end{table}

\subsection{Comparison with Video-based State-of-the-arts}
We compare our method with two video-based state-of-the-art stereo matching methods, i.e., DynamicStereo~\cite{b5} and BiDAStereo~\cite{b6}. Note that, existing video-based stereo matching methods are designed for fully supervised video-level training in nature scenes, requiring frame-wise annotations that are unavailable in our MIS scenarios. To enable a comparative analysis, we adapt these video-based methods by retraining them using the pseudo labels generated in our I2V stage. This comparison fails to fully capture the strength of TiS-TSL in learning robust temporal representations directly from sparse image-level annotations through the joint design of the I2V and V2V stages. The three methods are denoted as I2V+DynamicStereo, I2V+BiDAStereo, and I2V+V2V (Ours), respectively.

By testing on a single NVIDIA GeForce RTX 4090 GPU, we compare the performance and Run-time (ms) on SCARED (a resolution of 1024$\times$1280). The results are shown in Table \ref{tab:video_comparison}. As can be seen, benefiting from the spatio-temporal filtering of pseudo labels in the V2V stage, our method achieves the best performance. Furthermore, compared to the second-best video-based method, DynamicStereo, our method has a significantly lower Run-time (403 ms vs. 848 ms), demonstrating greater potential for clinical applications. 

\begin{figure}[htbp] 
\centering % 使图片在环境中水平居中
\includegraphics[width=0.48\textwidth]{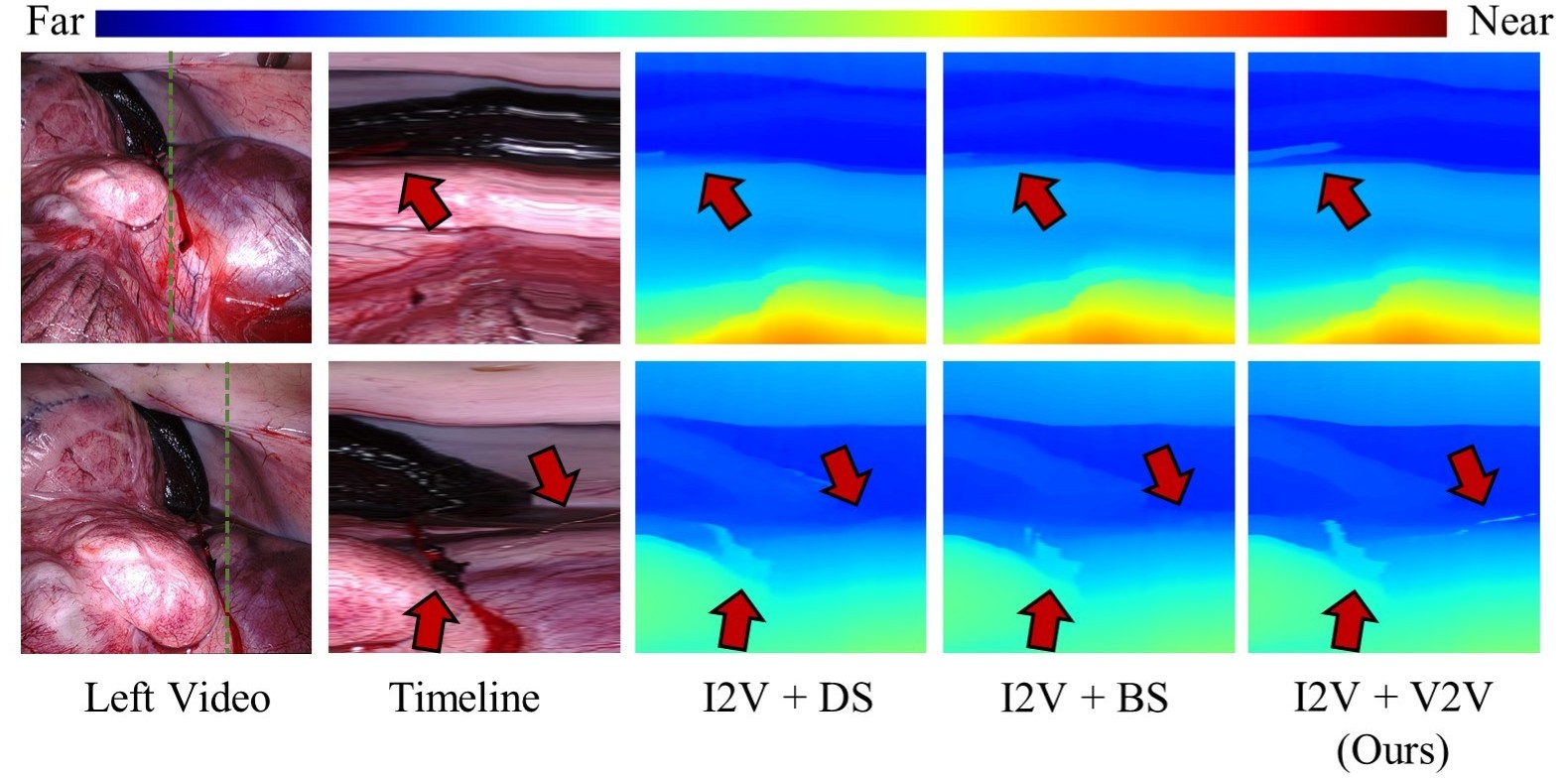} 

\caption{Qualitative comparisons with two video-based methods on SCARED videos. The second column represents the temporal image profiles
obtained by slicing the images along the timeline at green-line positions. The subsequent columns represent the corresponding disparity profiles. The red arrows indicate highlight regions. OpenCV Jet Colormap is used for visualization.} 
\label{fig:timeline video-based} % 图片标签，用于在文中引用

\end{figure}

Similar to Sec. \ref{sec:CIB}, we present two video cases as shown in Fig. \ref{fig:timeline video-based}. As can be seen, compared to other video-based methods, our method is capable of capturing fine and complicated structures, and the disparity profiles are also more consistent with the left videos, exhibiting the highest temporal consistency.

\subsection{Ablation Study}

To further verify the effectiveness of the two stages and two of their key components, we conduct ablation studies on SCARED. Moreover, we investigate the impact of the hyperparameters of ST-CFM.

\subsubsection{The effectiveness of the two stages and two of their key components} To verify the effectiveness of the two stages, i.e., I2V and V2V, and two of their key components, i.e., FVP and ST-CFM, we train four variants of TiS-TSL. The performance comparison results are shown in Table~\ref{tab:ablation_results_v2}. Furthermore, we treat the teacher model trained solely on labeled image data in the I2V stage as the baseline. 

As shown in the first three rows of Table~\ref{tab:ablation_results_v2}, utilizing unlabeled video data enhances the estimation performance of the student model. By incorporating temporal information processing through the architectural modification, i.e., FVP, we further reduce TEPE and EPE by 2.08\% and 4.37\%, respectively. The comparison between the $3^{rd}$ and $4^{th}$ row, video-level pseudo labels generated by the pre-trained teacher brings limited performance gains, especially in TEPE, which can be attributed to the lack of reliability assessment. From the last two rows of Table III, we can observe that ST-CFM brings a reduction in TEPE and EPE by 1.06\% and 3.08\%, respectively, which indicates that ST-CFM effectively suppresses erroneous information within pseudo labels.

\begin{table}[htbp] % 使用 'table' 环境确保表格占据单栏宽度
\centering % 居中表格
\caption{Ablation study of the two stages and two of their key components. Vid refers to training using unlabeled video data.} % 表格标题 (您可根据需要修改)
\label{tab:ablation_results_v2} % 表格标签，用于引用

\footnotesize % 减小字体大小，以适应单栏表格

% 列格式定义：
% l: 第一列 (Stage) 左对齐
% ccc ccc: 后面所有六列居中对齐
\begin{tabular}{*{1}{>{\centering\arraybackslash}m{0.4cm}} *{3}{>{\centering\arraybackslash}m{0.45cm}} *{1}{>{\centering\arraybackslash}m{1.0cm}} *{1}{>{\centering\arraybackslash}m{0.85cm}} *{1}{>{\centering\arraybackslash}m{1.1cm}} *{1}{>{\centering\arraybackslash}m{0.8cm}}}
    \toprule % 表格顶部粗线
    Stage & Vid & FVP & \shortstack{ST-\\CFM} & TEPE $\downarrow$ & $\delta^{3px}_t$ $\downarrow$ & TEPEr $\downarrow$ & EPE $\downarrow$ \\
    \midrule % 表格中间线

    % I2V 阶段的分组
    \multirow{3}{*}{I2V} & \ding{73} & $\times$ & $\times$ & 1.05 & 3.42 & 6.95 & 2.19 \\
    & \ding{72} & $\times$ & $\times$ & 0.96 & 2.90 & 5.54 & 2.06 \\
    & \ding{72} & $\checkmark$ & $\times$ & 0.94 & 2.80 & 5.25 & 1.97 \\
    \midrule % 分隔线

    % V2V 阶段的分组
    \multirow{2}{*}{V2V} & \ding{72} & $\checkmark$ & $\times$ & 0.94 & \textbf{2.78} & 5.20 & 1.95 \\
    & \ding{72} & $\checkmark$ & $\checkmark$ & \textbf{0.93} & \textbf{2.78} & \textbf{4.99} & \textbf{1.89} \\
    \bottomrule % 表格底部粗线
\end{tabular}
\end{table}

\subsubsection{Hyperparameters of ST-CFM} As illustrated in the soft threshold
mapping function Eq. \ref{eq:soft}, a higher $\epsilon$ value or a lower $\tau$ value makes the filtering mechanism more sensitive to inconsistent regions, consequently leading to lower tolerance; conversely, higher tolerance is observed. We further investigate the impact of these two hyperparameters on model performance, as shown in Table \ref{tab:epsilon_tau_ablation_simple}. Specifically, the model achieves optimal performance when $\epsilon$ and $\tau$ are set to 10 and 1, respectively. This demonstrates that the teacher model in the V2V stage has already developed a preliminary ability for temporal modeling. Therefore, a reduced tolerance for inconsistencies in the pseudo labels becomes necessary to provide more effective supervision for the student model.

\begin{table}[htbp] % 使用 'table' 环境确保表格占据单栏宽度
\centering % 居中表格
\caption{The ablation study on the hyperparameters $\epsilon$ and $\tau$ of ST-CFM.} 
\label{tab:epsilon_tau_ablation_simple} % 表格标签，用于引用

\small % 减小字体大小，以适应单栏表格

% 列格式定义：
% c: 所有列都居中对齐
\begin{tabular}{*2{>{\centering\arraybackslash}m{0.6cm}} *6{>{\centering\arraybackslash}m{1.2cm}}}
    \toprule % 表格顶部粗线
    \textbf{$\epsilon$} & \textbf{$\tau$} & TEPE $\downarrow$ & $\delta_t^{3px}$ $\downarrow$ & TEPEr $\downarrow$ & EPE $\downarrow$ \\
    \midrule % 表格中间线

    5 & 3 & 0.94 & 2.77 & 5.15 & 1.07 \\
    5 & 1 & 0.95 & 2.79 & 5.33 & 1.05 \\
    10 & 3 & 0.94 & 2.78 & 5.24 & 1.06 \\
    10 & 1 & \textbf{0.93} & \textbf{2.72} & \textbf{4.99} & \textbf{1.04} \\
    \bottomrule % 表格底部粗线
\end{tabular}
\end{table}

\section{CONCLUSION}
In this paper, we propose TiS-TSL, a novel time-switchable teacher-student learning framework for temporally consistent stereo matching in surgical videos under minimal supervision. At its core, the proposed time-switchable stereo model seamlessly integrates three operational modes, i.e., Image-Prediction (IP), Forward Video-Prediction (FVP), and Backward Video-Prediction (BVP), enabling flexible spatial refinement and bidirectional temporal propagation within a single model. The hierarchical training strategy progressively transfers knowledge from sparse image labels to unlabeled videos: the Image-to-Video (I2V) stage initializes temporal modeling through pseudo-supervision. The subsequent Video-to-Video (V2V) stage enforces cross-frame consistency via our proposed Spatio-Temporal Confidence Filtering Mechanism (ST-CFM), which identifies unreliable regions by evaluating forward-backward prediction agreement. Comprehensive validation on SCARED and Hamlyn demonstrates state-of-the-art performance. This work provides a practical solution for 3D surgical navigation systems where dense disparity annotations are clinically infeasible to acquire.

\section*{Acknowledgment}
	This work was supported in part by National Key R\&D Program of China (Grant No. 2023YFC2414900), Research grants from Wuhan United Imaging Surgical Co., Ltd.
    
% \section*{Acknowledgment}
	
% 	This work was supported in part by National Key R\&D Program of China (Grant No. 2023YFC2414900), Key R\&D Program of Hubei Province of China (No.2023BCB003), Fundamental Research Funds for the Central Universities (2021XXJS033), Research grants from Wuhan United Imaging Surgical Co., Ltd.

% \section*{Acknowledgment}
	
% 	This work was supported in part by National Key R\&D Program of China (Grant No. 2023YFC2414900), Key R\&D Program of Hubei Province of China (No.2023BCB003), Fundamental Research Funds for the Central Universities (2021XXJS033), Research grants from Wuhan United Imaging Surgical Co., Ltd.


\begin{thebibliography}{00}
\bibitem{b1} Y. Gao, Q. Wang, X. Rao, L. Xie and Y. Ying, ``OrangeStereo: A navel orange stereo matching network for 3D surface reconstruction,'' Comput. Electron. Agr., vol. 217, pp. 108626, February 2024.
\bibitem{b2} R. Wei, B. Li, H. Mo, B. Lu, Y. Long, B. Yang  \textit{et al}., ``Stereo dense scene reconstruction and accurate localization for learning-based navigation of laparoscope in minimally invasive surgery,'' IEEE Trans. Biomed. Eng., vol. 70, no. 2, pp. 488-500, February, 2023.
\bibitem{b3} X. Kang, M. Azizian, E. Wilson, K. Wu, A. Matrin, T. Kane \textit{et al}., ``Stereoscopic augmented reality for laparoscopic surgery,'' Surg. Endosc., vol. 28, pp. 2227-2235, February 2014.
\bibitem{b4} C.  Portalés, J. Gimeno, A. Salvador, A. García-Fadrique, and S. Casas-Yrurzum, ``Mixed Reality Annotation of Robotic-Assisted Surgery videos with real-time tracking and stereo matching,'' Comput. Graphics, vol. 110, pp. 125-140, February 2023.
\bibitem{b5} N. Karaev, I. Rocco, B.  Graham, N. Neverova, A. Vedaldi, and C. Rupprecht,``DynamicStereo: Consistent Dynamic Depth from Stereo Videos,'' in Proceedings of the IEEE/CVF Conference on Computer Vision and Pattern Recognition, 2023, pp. 13229-13239.
\bibitem{b6} J. Jing, Y. Mao, and K. Mikolajczyk, ``Match-stereo-videos: Bidirectional alignment for consistent dynamic stereo matching,'' in Proceedings of the European Conference on Computer Vision, 2024, pp. 415-432.
\bibitem{b7} H. Shi, Z. Wang, J. Lv, Y. Wang, P. Zhang, F. Zhu \textit{et al}., ``Semi-supervised learning via improved teacher-student network for robust 3D reconstruction of stereo endoscopic image,'' in Proceedings of the 29th ACM International Conference on Multimedia, 2021, pp. 4661-4669.
\bibitem{b8} Y. Zhou, S. He, H. Wang, F. Huang, M. Liu, Q. Li, and Z. Wang, ``Improved Self-supervised Monocular Endoscopic Depth Estimation based on Pose Alignment-friendly Dynamic View Selection,`` in 2024 IEEE International Conference on Bioinformatics and Biomedicine, IEEE, 2024, pp. 3005-3012.
\bibitem{b9} W. Cui, Y. Liu, Y. Li, M. Guo, Y. Li, X. Li \textit{et al}.,``Semi-supervised brain lesion segmentation with an adapted mean teacher model,'' in International Conference on Information Processing in Medical Imaging, Springer, 2019, pp. 554–565.
\bibitem{b10} L. Lipson, Z. Teed, and J. Deng,``RAFT-Stereo: Multilevel Recurrent Field Transforms for Stereo Matching,'' in 2021 International Conference on 3D Vision, IEEE, 2021, pp. 218-227.
\bibitem{b11} G. Xu, X. Wang, X. Ding, and X. Yang, ``Iterative geometry encoding volume for stereo matching,'' in Proceedings of the IEEE/CVF Conference on Computer Vision and Pattern Recognition, 2023, pp. 21919-21928.
\bibitem{b12} X. Wang, G. Xu, H. Jia, and X. Yang, ``Selective-stereo: Adaptive frequency information selection for stereo matching,'' in Proceedings of the IEEE/CVF Conference on Computer Vision and Pattern Recognition, 2024, pp. 19701-19710.
\bibitem{b13} Z. Teed, and J. Deng, ``RAFT: Recurrent All-Pairs Field Transforms for Optical Flow, '' in European conference on computer vision, Springer, 2020, pp. 402-419.
\bibitem{b14} Z. Li, W. Ye, D. Wang, F. Creighton, R. Taylor, G. Venkatesh \textit{et al}., ``Temporally Consistent Online Depth Estimation in Dynamic Scenes,'' in Proceedings of the IEEE/CVF Winter Conference on Applications of Computer Vision, 2023, pp. 3018-3027.
\bibitem{b15} A. Dosovitskiy, L. Beyer, A. Kolesnikov, D. Weissenborn, X. Zhai, T. Unterthiner \textit{et al}., ``An image is worth 16x16 words: Transformers for image recognition at scale,'' arXiv preprint arXiv:2010.11929, 2020.
\bibitem{b16} H. Shi, Z. Wang, Y. Zhou, D. Li, X. Yang, and Q. Li. ``Bidirectional semi-supervised dual-branch CNN for robust 3D reconstruction of stereo endoscopic images via adaptive cross and parallel supervisions,''IEEE Trans. Med. Imaging, vol. 42, no. 11, pp. 3269-3282, November 2023.
\bibitem{b17} S. Woo, J. Park, J. Lee, and I. S. Kewon. ``CBAM: Convolutional block attention module,'' in Proceedings of the European Conference on Computer Vision, 2018, pp. 3-19.
\bibitem{b18} J. Zhou, P. Ye, H. Zhang, J. Yuan, R. Qiang, L. Yang \textit{et al}., ``Consistency-aware Self-Training for Iterative-based Stereo Matching,'' in Proceedings of the IEEE/CVF Conference on Computer Vision and Pattern Recognition, 2025, pp. 16641-16650.
\bibitem{b19} I. Loshchilov, and F. Hutter.,``Decoupled weight decay regularization,'' arXiv preprint arXiv:1711.05101, 2017.
\bibitem{b20} D. Recasens, J. Lamarca, J. M. Fácil, J. Montiel, and J. Civera, “Endodepth-and-motion: Reconstruction and tracking in endoscopic videos using depth networks and photometric constraints,” IEEE Robot. Autom. Lett., vol. 6, no. 4, pp. 7225–7232, 2021.
\bibitem{b21} J. Chang, and Y. Chen, “Pyramid stereo matching network,” in Proceedings of the IEEE/CVF Conference on Computer Vision and Pattern Recognition, 2018, pp. 5410-5418.
\bibitem{b22} X. Guo, K. Yang, X. Wang, and H. Li, “Group-wise Correlation Stereo Network,” in Proceedings of the IEEE/CVF conference on computer vision and pattern recognition, 2019, pp. 3273-3282.
\bibitem{b23} G. Xu, J. Cheng, P. Guo, and X. Yang, “Attention Concatenation Volume for Accurate and Efficient Stereo Matching,” in Proceedings of the IEEE/CVF Conference on Computer Vision and Pattern Recognition, 2022, pp. 12981-12990.
\bibitem{b24} B. Wen, M. Trepte, J. Aribido, J. Kautz, O. Gallo, and S. Birchfield, “FoundationStereo: Zero-Shot Stereo Matching,” in Proceedings of the IEEE/CVF Conference on Computer Vision and Pattern Recognition, 2025, pp. 5249-5260.
\end{thebibliography}
\end{document}